\documentclass[12pt]{article}
\pdfoutput=1 
\PassOptionsToPackage{numbers}{natbib}
\usepackage[preprint]{neurips_2024}

\usepackage[utf8]{inputenc} 
\usepackage[T1]{fontenc}    
\usepackage{hyperref}       
\usepackage{url}            
\usepackage{booktabs}       
\usepackage{amsfonts}       
\usepackage{amsmath}        
\usepackage{graphicx}       
\usepackage{float}          
\usepackage{multirow}

\usepackage{listings}
\usepackage{xcolor}
\title{Litespark Inference For CPUs:\\Ultra-Fast SIMD Framework for Ternary (1.58-bit) Language Models}

\author{Nii Osae Osae Dade
$\quad$ Tony Morri
$\quad$ Moinul Hossain Rahat\textsuperscript{*}\\
\\
\textbf{$\quad$ Sayandip Pal
$\quad$ Rickston Pinto\textsuperscript{*}}\\
\\
Mindbeam AI\textsuperscript{$\dagger$} \\
}

\begin{document}

\maketitle

\begingroup
\renewcommand{\thefootnote}{\fnsymbol{footnote}}
\footnotetext[1]{Work done at Mindbeam AI}
\footnotetext[2]{Correspondence to: research@mindbeam.ai}
\endgroup

\begin{abstract}
Large language models (LLMs) have transformed artificial intelligence, but their computational requirements remain prohibitive for most users. Standard inference demands expensive datacenter GPUs or cloud API access, leaving over one billion personal computers underutilized for AI workloads. Ternary models offer a path forward: their weights are constrained to $\{-1, 0, +1\}$, theoretically eliminating the need for floating-point multiplication. However, existing frameworks fail to exploit this structure, treating ternary models as dense floating-point networks. We address this gap with custom SIMD kernels that replace matrix multiplication with simple addition and subtraction operations, targeting the integer dot product instructions available on modern CPUs. Our implementation, \texttt{Litespark-Inference}, is pip-installable and integrates directly with HuggingFace, achieving 18.15$\times$ higher throughput, 7.15$\times$ faster time-to-first-token and 6.03$\times$ memory reduction compared to standard PyTorch inference on Apple Silicon, with comparable or higher throughput speedups up to 95.81$\times$ on Intel and AMD processors.
\end{abstract}

\section{Introduction}

Large language models have demonstrated remarkable capabilities across a wide range of tasks, from natural language understanding to code generation. However, deploying these models remains challenging due to their substantial computational and memory requirements. GPU-based inference requires datacenter GPUs costing \$25,000--\$40,000 per unit \cite{h100pricing}, or cloud API access at \$2.50--\$10 per million tokens \cite{openaipricing}. Meanwhile, there are over one billion personal computers in use worldwide \cite{pcstats}. These devices have capable CPUs that remain underutilized for AI workloads, and could run LLMs if the software existed to exploit them.

Beyond cost, on-device inference offers compelling advantages for privacy and accessibility. Running models locally keeps sensitive data on the user's machine, avoiding transmission to external servers. It enables offline operation without network connectivity, reduces latency by eliminating round-trip communication, and democratizes access to AI capabilities for users in regions with limited cloud infrastructure or unreliable internet.

Recent advances in model quantization have reduced the resource requirements for LLM inference. Approaches like 4-bit and 8-bit quantization \cite{llama-cpp} have made it possible to run smaller models on consumer hardware. Ternary models, e.g., BitNet \cite{bitnet}, TriLM \cite{TriLM}, take quantization to its theoretical limit: ternary weights that can only take values from $\{-1, 0, +1\}$. This extreme quantization has a profound implication: matrix multiplication, the dominant operation in neural networks, becomes unnecessary. When weights are ternary, computing $y = \sum_j x_j \cdot w_j$ reduces to simple conditional operations, adding $x_j$ to the accumulator when $w_j = +1$, subtracting $x_j$ when $w_j = -1$, and skipping the operation entirely when $w_j = 0$.


While the theoretical advantages of ternary models are clear, their practical CPU-based inference
has only recently become feasible. Three developments make efficient CPU inference for these models timely. 

First, the required hardware capabilities have only recently become widespread.
The SIMD instructions required for efficient integer dot products
arrived when
Intel introduced AVX-512 VNNI with Ice Lake (2019), ARM Holdings developed and deployed SDOT as part of their NEON Advanced SIMD architecture, first appearing in Apple's M1 chip (2020) and now ubiquitous across billions of ARM-based mobile devices, and AMD brought AVX-512 to consumer chips with Zen4 (2022). ARM's foresight in incorporating efficient integer dot product capabilities into their architecture has proven particularly impactful, as ARM-based processors now power not only Apple's entire Mac lineup but also the vast majority of smartphones and tablets worldwide. Before these instructions, CPU-based neural network inference was bottlenecked by the lack of efficient integer arithmetic.

Second, production-quality ternary LLMs have become available. While ternary
quantization has been studied theoretically, Microsoft's release of BitNet b1.58 in 2024 provided a
2-billion parameter model trained on 4 trillion tokens that demonstrates competitive performance
with full-precision models of similar size.

Third, the demand-supply gap has widened. LLM usage is
growing rapidly, but GPU availability remains constrained and costs remain high. Most developers and researchers cannot justify dedicated GPU hardware for experimentation. Efficient CPU inference offers an accessible alternative that runs on hardware already owned.

Given these converging developments in hardware, models, and demand, we address the gap between
the theoretical potential of ternary models and its practical realization.
Despite the theoretical advantage of ternary weights, existing inference frameworks do not fully exploit their structure. Standard PyTorch \cite{pytorch} treats these models as dense floating-point networks, missing the opportunity for optimization. Specialized frameworks like llama.cpp focus on 4-bit quantization rather than ternary operations.

We present custom SIMD kernels that 
eliminate multiplication for ternary inference. Our implementation targets three major CPU architectures. On Apple Silicon (M1--M5), we use NEON SDOT instructions with 128-bit vectors. On Intel Ice Lake and AMD Zen4 (and later), we use AVX-512 VNNI instructions with 512-bit vectors. On Intel Core Ultra 9, we use AVX-VNNI instructions with 256-bit vectors.

This work makes three contributions. First, we develop custom SIMD kernels that exploit ternary weight structure for multiplication-free inference, replacing floating-point matrix multiplication with integer addition and subtraction via hardware dot product instructions. Second, we package these kernels as a pip-installable Python library called \texttt{Litespark-Inference}, with automatic platform detection and HuggingFace Transformers \cite{huggingface} integration. Third, we provide comprehensive benchmarks for the specific case of BitNet b1.58 model, demonstrating 18.15--97.46$\times$ throughput improvement and approximately 6$\times$ memory reduction across platforms.

\section{Background}

\subsection{Neural Network Quantization}

Neural networks store their learned knowledge in \textit{weights}, numerical parameters that transform input data into predictions. In standard networks, weights are stored as 32-bit floating-point numbers (float32), allowing fine-grained values like 0.00142 or -3.7891. However, this precision comes at a cost: a 2-billion parameter model requires 8 GB of memory just for weights.

\textit{Quantization} reduces memory by storing weights with fewer bits. Common approaches range from 16-bit formats like bfloat16 and float16, which halve memory with minimal accuracy loss, down through 8-bit integer representation (one quarter the memory with small accuracy loss) and 4-bit quantization (one-eighth the memory with moderate accuracy loss). At the extreme end lies ternary quantization at 1.58 bits, where weights are restricted to only three possible values: $\{-1, 0, +1\}$.

The idea of extreme weight quantization has a rich history. BinaryConnect \cite{binaryconnect} first demonstrated that neural networks could be trained with a set of binary weights $\{-1, +1\}$ during forward and backward propagation, achieving competitive results on MNIST and CIFAR-10. This work established that the computational benefits of binary weights, replacing multiplications with simple sign changes, could be realized without catastrophic accuracy loss, laying the groundwork for modern ternary approaches.

Recent work has developed increasingly sophisticated quantization techniques for large language models. LLM.int8() \cite{llmint8} introduced mixed-precision decomposition to handle outlier features in billion-scale transformers, enabling 8-bit inference without performance degradation. GPTQ \cite{gptq} introduced accurate post-training quantization using approximate second-order information, enabling 3-4 bit quantization with minimal accuracy loss. AWQ \cite{awq} demonstrated that protecting only 1\% of salient weights significantly reduces quantization error
. SmoothQuant \cite{smoothquant} enables W8A8 quantization by migrating quantization difficulty from activations to weights. QLoRA \cite{qlora}, building on LoRA's \cite{lora} low-rank adaptation approach, extended these techniques to enable efficient fine-tuning of 4-bit quantized models on consumer hardware. For a comprehensive overview of quantization methods, we refer the readers to recent surveys \cite{quantsurvey}.

Ternary quantization is the most extreme form. Each weight can only say ``add,'' ``subtract,'' or ``skip'', nothing in between. This seems limiting, but models like BitNet and TriLM trained natively with this constraint demonstrate that ternary quantized models can match the performance of full-precision models.

\subsection{Ternary Model Architecture}

Ternary models follow the standard transformer architecture \cite{vaswani2017attention} but constrain all linear layer weights to ternary values. For our benchmarks, we use BitNet b1.58 \cite{bitnet158}, a 2-billion parameter model trained on 4 trillion tokens that demonstrates competitive performance with full-precision counterparts.

\subsubsection{The Computational Advantage}

In a standard linear layer, the forward pass computes:
\begin{equation}
    Y = XW
\end{equation}
where $X \in \mathbb{R}^{M \times K}$ is the input activation matrix ($M$ tokens, $K$ features) and $W \in \mathbb{R}^{K \times N}$ is the weight matrix. Each output element requires $K$ multiply-accumulate operations:
\begin{equation}
    Y_{ij} = \sum_{k=1}^{K} X_{ik} \cdot W_{kj}
\end{equation}

For a typical layer with $K = 2048$ and $N = 8192$, this means 16 million multiplications per token, and a 2B parameter model has dozens of such layers.

With ternary weights $W \in \{-1, 0, +1\}^{K \times N}$, multiplication becomes trivial:
\begin{equation}
    Y_{ij} = \sum_{k: W_{kj}=+1} X_{ik} - \sum_{k: W_{kj}=-1} X_{ik}
\end{equation}

The computation reduces to addition and subtraction. Weights equal to zero contribute nothing and can be skipped entirely.

\subsubsection{Memory Savings}

The memory advantage of ternary weights is straightforward to quantify. Since each weight takes one of three values, two bits per parameter suffice to encode the full weight space. This stands in stark contrast to the 32 bits required for standard float32 representation.

For a 2B parameter ternary model, this difference translates to substantial memory savings. A float32 representation would require $2 \times 10^9 \times 32$ bits, or approximately 8 GB of memory for weights alone. A pure 2-bit ternary encoding would theoretically need only $2 \times 10^9 \times 2$ bits, approximately 500 MB.

In practice, our implementation uses 8 bits per weight rather than the theoretical minimum of 2 bits. This decision prioritizes compatibility with hardware dot product instructions, which expect 8-bit integer inputs. The practical storage footprint is approximately 556 MB after accounting for alignment padding, still achieving a 14$\times$ reduction compared to float32. We discuss the rationale for this design choice in detail in Section 4.1.1.

\subsection{SIMD Instructions on Modern CPUs}

While ternary weights reduce the computational complexity of neural network operations, realizing
these benefits in practice requires specialized hardware support. Modern CPUs provide such support
through SIMD instructions.

\textit{Single Instruction Multiple Data} (SIMD) is a CPU feature that processes multiple values simultaneously with a single instruction. Rather than adding two numbers one at a time, a SIMD instruction can add 16 or 64 pairs of numbers in one cycle.

Modern CPUs have wide \textit{vector registers} that hold multiple data elements. A 128-bit register can hold 16 int8 values (or 4 int32 values), a 256-bit register holds 32 int8 values, and a 512-bit register holds 64 int8 values.

\subsubsection{Dot Product Instructions}

The critical enabler for efficient ternary inference is a family of specialized SIMD instructions designed for neural network workloads. These instructions compute integer dot products with native hardware support, eliminating the need for explicit loops over vector elements.

ARM's NEON instruction set, available on Apple Silicon (M1 through M5) and mobile ARM
processors, provides the SDOT instruction. ARM Holdings developed the NEON Advanced SIMD architecture as a key component of their energy-efficient processor designs. The SDOT (signed dot product) instruction, part of ARM's ARMv8.2-A DotProd extension, computes the dot product of 16 int8 values, accumulating into four int32 results. Each instruction performs 16 multiply-add operations with exceptional power efficiency. This instruction set is now deployed across billions of devices worldwide, from Apple's Mac lineup to Android smartphones and embedded systems. ARM's comprehensive documentation \cite{armneon} provides detailed specifications for developers leveraging these capabilities. The widespread adoption of ARM's NEON architecture has made efficient on-device AI inference accessible to a global user base.

On x86 architectures, Intel and AMD provide the AVX-512 VNNI instruction set (Intel Ice Lake and
later, AMD Zen4 and later). The VPDPBUSD instruction computes dot products across 64 int8 values per instruction, 4$\times$ the throughput of NEON. Intel provides detailed documentation as part of Intel Deep Learning Boost \cite{intelvnni}.

For Intel processors without full AVX-512 support, the AVX-VNNI instruction set (available starting
with 12th generation Core processors and Core Ultra) offers a middle ground. These 256-bit
instructions process 32 int8 values per operation, providing better throughput than NEON while
maintaining compatibility with processors that lack full AVX-512 capability.

Table \ref{tab:simd_comparison} summarizes the SIMD capabilities across platforms.

\begin{table}[h]
\centering
\begin{tabular}{lccc}
\toprule
\textbf{Platform} & \textbf{Vector Width} & \textbf{int8 ops/instruction} & \textbf{Instruction} \\
\midrule
Apple M1--M5 & 128-bit & 16 & SDOT \\
Intel Core Ultra & 256-bit & 32 & VPDPBUSD \\
Intel Ice Lake+ & 512-bit & 64 & VPDPBUSD \\
AMD Zen4+ & 512-bit & 64 & VPDPBUSD \\
\bottomrule\\
\end{tabular}

\caption{SIMD capabilities for integer dot products across CPU platforms.}
\label{tab:simd_comparison}
\end{table}

\section{Related Work}

With an understanding of the hardware capabilities available, we now examine how existing frameworks
approach CPU inference for LLMs and where our work fits within this landscape. Several strategies
for efficient CPU-based inference have been explored, each with distinct architectural choices and
performance tradeoffs. Table~\ref{tab:related_work} provides a comparative overview of the major frameworks.

\begin{table}[h]
\centering
\begin{tabular}{lcccc}
\toprule
\textbf{Framework} & \textbf{Precision} & \textbf{Ternary-Optimized} & \textbf{pip install} & \textbf{HuggingFace} \\
\midrule
llama.cpp \cite{llama-cpp} & 4/8-bit & No & No & No \\
T-MAC \cite{tmac} & Ternary & Yes (LUT) & No & No \\
BitNet.cpp \cite{bitnetcpp} & Ternary & Yes & No & No \\
\textbf{Litespark-Inference} & Ternary & Yes (SIMD) & Yes & Yes \\
\bottomrule\\
\end{tabular}
\caption{Comparison of CPU inference frameworks.}
\label{tab:related_work}
\end{table}

The llama.cpp and GGML frameworks \cite{llama-cpp} have established themselves as widely-used tools for CPU inference of quantized language models. These frameworks primarily target 4-bit and 8-bit quantization schemes, applying general-purpose quantization techniques that do not exploit the specific algebraic properties of ternary weights. While effective for their target precision levels, these approaches cannot leverage the multiplication-free computation enabled by ternary quantization.

T-MAC \cite{tmac} takes a fundamentally different approach to ternary inference through lookup-table-based (LUT) computation. Rather than computing dot products directly, T-MAC precomputes results for all possible input combinations and retrieves them during inference via table lookups. This method can achieve strong performance by trading computation for memory bandwidth. However, the approach requires substantial memory to store the lookup tables and introduces implementation complexity in managing table organization and access patterns.

Microsoft's BitNet.cpp \cite{bitnetcpp} provides an official reference implementation for ternary inference with specialized optimizations. While this framework demonstrates effective ternary inference, it requires specific build configurations, compiler flags, and manual setup. The lack of integration with standard Python package managers and the HuggingFace ecosystem creates friction for practitioners seeking to experiment with ternary models.

\text{FlashAttention} \cite{flashattention} addresses a different bottleneck, the quadratic memory complexity of attention, through IO-aware algorithms that reduce memory reads/writes between GPU HBM and SRAM. While orthogonal to our work on linear layer optimization, FlashAttention demonstrates the importance of hardware-aware algorithm design for efficient inference.

Litespark-Inference differs by using direct SIMD dot product instructions rather than lookup tables, resulting in simpler code that is easier to understand and maintain. The implementation is packaged as a standard pip-installable library with automatic HuggingFace model loading.

\section{Method}

Having established the theoretical foundations and limitations of existing approaches, we now describe
our implementation. The key challenge is bridging the gap between the theoretical simplicity of
ternary operations (just add and subtract) and the practical requirements of hardware execution,
which must account for memory alignment constraints, numerical precision through quantization, and
computational stability.

\subsection{Kernel Design}

Our kernel design philosophy centers on three core principles. First, we exploit the ternary weight structure to eliminate multiplication operations entirely, replacing them with conditional addition and subtraction. Second, we leverage hardware SIMD dot product instructions to achieve data-level parallelism, processing multiple elements simultaneously. Third, we maintain numerical accuracy through careful quantization of activations and systematic correction of quantization-induced biases. The following subsections detail how each principle is realized in our implementation.

\subsubsection{Weight Representation}

A natural question is how to store ternary weights. Since each weight can take one of the three possible values from the set $\{-1$, $0$, $+1\}$, two bits per weight would suffice. However, we store weights as 8-bit signed integers instead.

This choice is driven by hardware constraints. The SIMD dot product instructions we rely on (SDOT, VPDPBUSD) expect 8-bit integer inputs. Using 2-bit packing would require unpacking weights before every computation, negating the performance benefit. By storing weights as int8, we can feed them directly to hardware instructions with no preprocessing.

The memory cost of this choice is modest. For a 2B parameter ternary model, 2-bit packing would require approximately 500 MB, while our 8-bit storage requires approximately 556 MB after alignment padding. The 56 MB difference is small compared to the 14$\times$ reduction from float32 (8 GB), and the performance gain from direct hardware utilization far outweighs this cost.

\subsubsection{Activation Quantization}

While weights are ternary, input activations (the values flowing through the network) are continuous floating-point numbers. To use integer dot product instructions, we must quantize these activations to int8.

We employ \textit{per-row symmetric quantization}. For each row of the activation matrix (corresponding to one token), we find the maximum absolute value and scale all values to fit within the int8 range $[-127, +127]$:
\begin{equation}
    x_{\text{int8}} = \text{round}\left(\frac{x \cdot 127}{\max(|x|)}\right), \quad \text{scale} = \frac{\max(|x|)}{127}
\end{equation}

The scale factor is saved so we can convert results back to float32 after computation. This approach preserves relative magnitudes within each token's activations, if one value was twice another before quantization, it remains approximately twice as large after.

\subsubsection{The Computation Pipeline}

The complete forward pass through a linear layer proceeds through four stages. First, activations are quantized by converting float32 inputs to int8 while saving the scale factor. Next, the kernel computes dot products using SIMD instructions to calculate $\sum_k x_k \cdot w_k$ for each output. The results then undergo zero-point correction, subtracting the bias introduced by quantization (explained below). Finally, the int32 results are rescaled back to float32 by multiplying with the saved scale factor.

\subsubsection{SIMD Dot Product}

The core computation uses hardware dot product instructions that process multiple int8 pairs simultaneously. On ARM-based processors, ARM's NEON \texttt{vsdotq\_s32} instruction computes the dot product of 16 int8 value pairs, accumulating into 4 int32 results. ARM's design philosophy emphasizes power efficiency alongside performance, making these instructions particularly well-suited for mobile and edge deployment scenarios. On Intel and AMD processors with AVX-512, the \texttt{\_mm512\_dpbusd\_epi32} instruction processes 64 int8 pairs per instruction, while the AVX-VNNI variant \texttt{\_mm256\_dpbusd\_epi32} handles 32 int8 pairs per instruction on Intel Core Ultra processors.

These instructions are designed for neural network inference and achieve high throughput on modern processors, with multiple operations completing per cycle when pipelined. The int32 accumulator prevents overflow even when summing thousands of int8 products.

\subsubsection{Zero-Point Correction}

Symmetric int8 quantization maps floating-point zero to integer zero. However, the quantization formula can introduce a small bias when values are not perfectly centered. This bias, called the \textit{zero-point}, must be corrected to maintain accuracy.

The correction is straightforward. We precompute the sum of each weight column:
\begin{equation}
    \text{col\_sum}_j = \sum_i W_{ij}
\end{equation}

After the dot product, we subtract $\text{zero\_point} \times \text{col\_sum}_j$ from each output $j$. Since column sums are constant for a given model, they are computed once during model loading and reused for every inference.

\subsection{Platform-Specific Implementations}

The diversity of SIMD instruction sets across modern CPU architectures necessitates platform-specific
kernel implementations. While the algorithmic approach remains consistent across platforms, each
target requires its own low-level implementation tailored to the available instruction set and register
organization. Table \ref{tab:platforms} summarizes the configurations.
\begin{table}[h]
\centering
\begin{tabular}{llcc}
\toprule
\textbf{Platform} & \textbf{Instruction} & \textbf{Vector Width} & \textbf{Alignment} \\
\midrule
Apple Silicon (M1--M5) & NEON SDOT & 128-bit & 16 bytes \\
Intel Ice Lake+ & AVX-512 VNNI & 512-bit & 64 bytes \\
AMD Zen4+ & AVX-512 VNNI & 512-bit & 64 bytes \\
Intel Core Ultra & AVX-VNNI & 256-bit & 32 bytes \\
\bottomrule\\
\end{tabular}
\caption{Platform-specific kernel configurations. Alignment refers to the memory boundary weights must be padded to for optimal SIMD access.}
\label{tab:platforms}
\end{table}

\text{Memory alignment} is critical for SIMD performance. Vector load instructions work most efficiently when data starts at addresses divisible by the vector width (i.e., addresses that are multiples of 16, 32, or 64 bytes depending on the platform). Misaligned loads
may require multiple memory transactions or trigger slower unaligned load paths in the hardware.

To ensure optimal performance, we pad weight matrices to the appropriate boundary during model loading such that each row
begins at an address satisfying the platform's alignment requirement. This padding introduces negligible memory overhead but ensures every load operation runs at full speed.

Our implementation incorporates automatic platform detection to select the appropriate kernel at
runtime. During library initialization, the system queries the CPU's feature flags (using CPUID on
x86 or equivalent mechanisms on ARM) to determine which SIMD instruction sets are available.
Based on this information, the library dispatches to the corresponding platform-specific kernel. This
design eliminates the need for users to manually specify their hardware configuration, simplifying
deployment across heterogeneous environments.

\subsection{Numerical Accuracy}

Beyond performance optimization, another practical consideration for deployment is the tradeoff
between computational efficiency and numerical precision. Integer quantization introduces small numerical differences compared to full float32 computation. For most applications, these differences are imperceptible, the model generates the same text.

Our inference path uses int8 quantized activations for maximum speed. When limiting precision to int8, numerical deviations can occur. We quantify this using the "maximum logit difference," defined as the largest deviation from the reference output across all logits. Our measurements show this difference is approximately 0.68 compared to float32 reference computation. In practice, this small deviation has no observable effect on generation quality: the same tokens are sampled, and the output text remains indistinguishable.

The same quantized path is used across Apple Silicon, Intel, and AMD platforms, allowing the implementation to preserve a consistent accuracy-performance tradeoff while selecting the appropriate SIMD kernel for each CPU architecture.

\section{Implementation}

The previous section described the algorithmic techniques underlying our kernels. This section turns to engineering concerns: how we package these kernels into a usable system that developers can adopt without specialized knowledge.

A key goal of this work is accessibility. Many efficient inference implementations require complex build processes, specific compiler versions, or manual hardware configuration. We designed our implementation to be installable with a single command and usable without any knowledge of the underlying SIMD optimizations. Our implementation is available at \url{https://github.com/Mindbeam-AI/Litespark-Inference}.

\subsection{System Architecture}

The implementation has four main components:

\textbf{C++ SIMD kernels.} The performance-critical dot product operations are implemented in C++ using platform-specific intrinsics (NEON for ARM-based processors, AVX-512/AVX-VNNI for x86). These kernels are compiled as PyTorch C++ extensions, allowing them to be called directly from Python with minimal overhead. Each platform has its own kernel file, and the build system automatically compiles only the relevant one.

\textbf{Python interface.} The user-facing API is pure Python, handling model loading, tokenization, and generation. Users interact with familiar PyTorch-style methods (\texttt{model.generate()}, \texttt{model.forward()}) without needing to know that custom kernels are running underneath.

\textbf{Automatic platform detection.} The runtime dispatching mechanism described in Section 4 is implemented by querying CPU feature flags at import time. The library then loads the appropriate pre-compiled kernel, making the optimization transparent to users.

\textbf{KV cache.} Language models generate text one token at a time. Without caching, each new token would require recomputing attention over the entire sequence, an $O(n^2)$ cost that grows quadratically with sequence length. The KV (key-value) cache stores intermediate attention results, reducing each generation step to $O(n)$. Our implementation uses a pre-allocated cache to avoid memory allocation during generation.

\subsection{Weight Conversion}

Ternary models distributed via HuggingFace store weights in standard floating-point format to ensure compatibility with existing frameworks. Our implementation performs an offline conversion to the optimized int8 ternary format described in Section 4.1.1.

The conversion process proceeds through several stages. The system first downloads the model from HuggingFace, with subsequent downloads served from a local cache. Each weight is then rounded to the nearest ternary value ($-1$, $0$, or $+1$) and converted to int8 representation. The weight matrices are padded to satisfy SIMD alignment requirements for the target platform. During this process, we also precompute the column sums required for zero-point correction (Section 4.1.5). The converted model is serialized to disk, enabling fast loading on subsequent runs.

This conversion overhead is incurred only on the first load of a model. All subsequent loads directly deserialize the converted representation, eliminating the conversion latency.
\subsection{User Interface}

The package provides both Python and command-line interfaces. The Python API integrates with HuggingFace, automatically downloading and converting ternary models:

\begin{lstlisting}[language=Python]
from litespark_inference import load_model

model, tokenizer = load_model("bitnet-2b")
input_ids = tokenizer.encode("Hello, world!", return_tensors="pt")
output = model.generate(input_ids, max_new_tokens=100)
print(tokenizer.decode(output[0]))
\end{lstlisting}

For users who prefer the command line, we provide a CLI with three modes:

\begin{lstlisting}[language=bash]
litespark-inference generate "The meaning of life is"
litespark-inference chat
litespark-inference benchmark
\end{lstlisting}

The \texttt{generate} command produces a single completion. The \texttt{chat} command starts an interactive session with conversation history. The \texttt{benchmark} command measures throughput and memory usage on the current hardware.

\subsection{Installation}

Installation requires only pip:

\begin{lstlisting}[language=bash]
pip install litespark-inference
\end{lstlisting}

\subsection{Reproducibility}

All code, benchmarking scripts, and instructions for reproducing our results are available in the repository \href{https://github.com/Mindbeam-AI/Litespark-Inference}{\texttt{Litespark-Inference}}. Experiments were run with Python 3.12 and PyTorch 2.11.0. The benchmark command included in the package measures throughput and memory usage on the user's hardware, enabling direct comparison with our reported results. Model weights are automatically downloaded from HuggingFace on first use.

\section{Experimental Results}

\subsection{Setup}

We evaluate our implementation using the Microsoft BitNet b1.58 2B-4T model, a 2-billion parameter ternary model trained on 4 trillion tokens. This was the only publicly available ternary model of significant scale at the time this research was conducted.

We tested on three hardware platforms representing the major CPU architectures in consumer devices today: Apple M5 Max (MacBook Pro), Intel Ice Lake and AMD Zen4 processors with AVX-512 VNNI support, and Intel Core Ultra 9 with AVX-VNNI.

We establish PyTorch with the native dense CPU backend as our baseline. This configuration represents the standard approach of running ternary models through conventional dense
matrix operations, treating ternary weights as regular floating-point values.

We report three metrics. First, memory consumption is measured as peak RSS (Resident
Set Size) during inference, capturing the actual physical memory footprint of the process. Second,
TTFT (time to first token) measures the latency from prompt submission to first token generation,
characterizing system responsiveness. Third, throughput is measured as tokens generated per second
during autoregressive generation with KV caching enabled, characterizing the sustained generation
rate for complete responses.

\subsection{Apple Silicon Results}

\begin{figure}[H]
\centering
\vspace*{0.5cm}
\includegraphics[width=\textwidth]{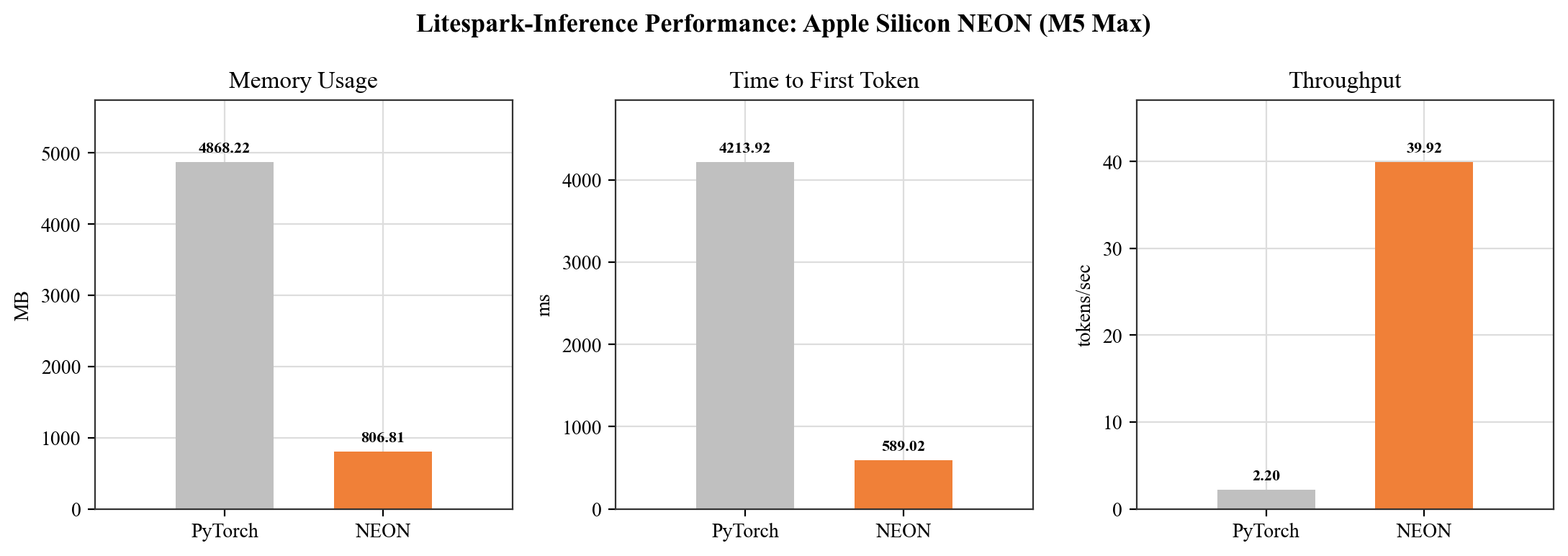}
\caption{Performance comparison on Apple Silicon NEON (M5 Max). Litespark-Inference achieves 6.03$\times$ memory reduction, 7.15$\times$ faster TTFT and 18.15$\times$ higher throughput compared to PyTorch.}
\label{fig:apple_results}
\end{figure}

Figure \ref{fig:apple_results} and Table \ref{tab:apple_results} show the performance comparison on Apple Silicon NEON (M5 Max). We compare standard PyTorch against Litespark-Inference using the NEON SDOT kernel with int8 quantized activations.

\begin{table}[h]
\centering
\begin{tabular}{lccc}
\toprule
\textbf{Metric} & \textbf{PyTorch} & \textbf{NEON} & \textbf{Improvement} \\
\midrule
Memory (MB) & 4,868.22 & 806.81 & 6.03$\times$ \\
TTFT (ms) & 4,213.92 & 589.02 & 7.15$\times$ \\
Throughput (tok/s) & 2.20 & 39.92 & 18.15$\times$ \\
\bottomrule\\
\end{tabular}
\caption{Detailed results on Apple Silicon NEON (M5 Max). Memory is peak RSS, TTFT is time to first token, throughput is tokens per second during generation.}
\label{tab:apple_results}
\end{table}

Litespark-Inference achieves substantial improvements in memory usage, time to first token, and sustained throughput. Memory usage drops by 6.03$\times$, from 4,868.22 MB to 806.81 MB. This allows the model to fit comfortably in RAM on devices with 8 GB of memory. Time to first token improves by 7.15$\times$, from 4,213.92 ms to 589.02 ms. Throughput increases by 18.15$\times$, from 2.20 tokens per second to 39.92 tokens per second. At nearly 40 tokens per second, text appears instantly; at 2.20 tokens per second, users would wait noticeably longer for complete responses.

\subsection{Intel and AMD Results}

We evaluate on two x86 configurations: processors with full AVX-512 VNNI support (512-bit vectors), and Intel Core Ultra 9 with AVX-VNNI (256-bit vectors).

\begin{figure}[H]
\centering
\vspace*{0.5cm}
\includegraphics[width=\textwidth]{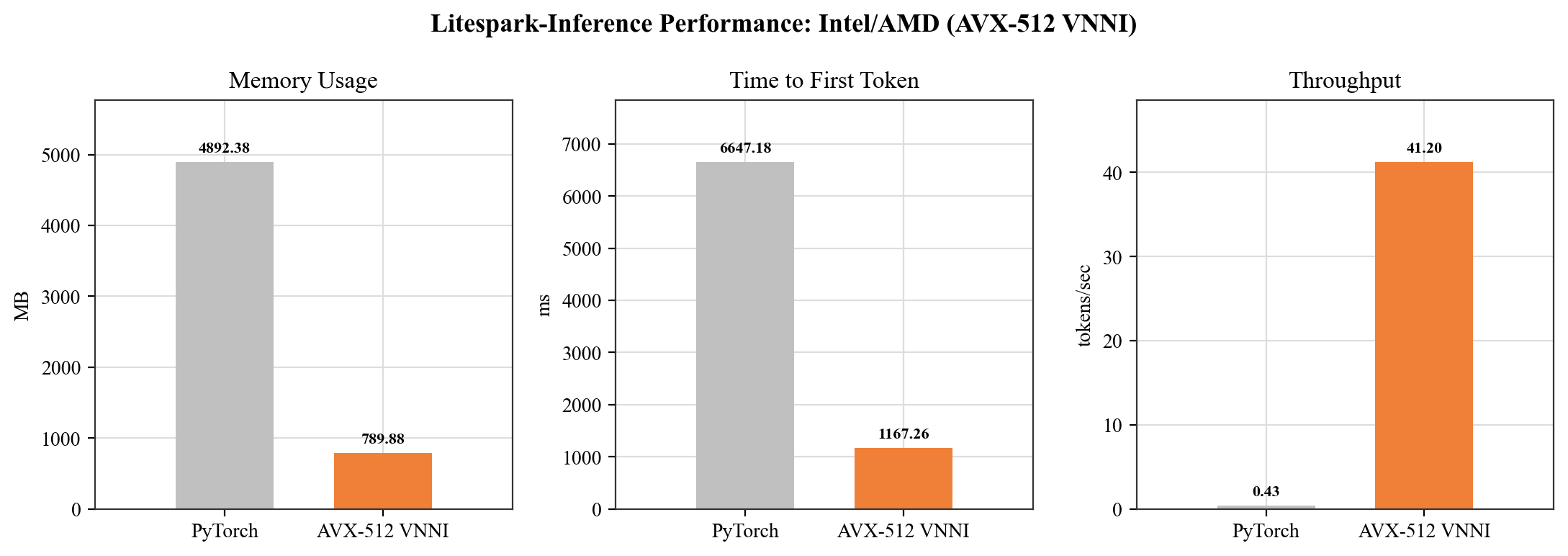}
\caption{Performance comparison on Intel/AMD (AVX-512 VNNI).}
\label{fig:avx512_results}
\end{figure}

\begin{table}[h]
\centering
\begin{tabular}{lccc}
\toprule
\textbf{Metric} & \textbf{PyTorch} & \textbf{AVX-512 VNNI} & \textbf{Improvement} \\
\midrule
Memory (MB) & 4,892.38 & 789.88 & 6.19$\times$ \\
TTFT (ms) & 6,647.18 & 1,167.26 & 5.69$\times$ \\
Throughput (tok/s) & 0.43 & 41.20 & 95.81$\times$ \\
\bottomrule\\
\end{tabular}
\caption{Results on Intel Ice Lake / AMD Zen4 (AVX-512 VNNI).}
\label{tab:avx512_results}
\end{table}

The AVX-512 VNNI kernel achieves 41.20 tokens per second with a 95.81$\times$ speedup. The 512-bit vector width allows processing 64 int8 values per instruction, compared to 16 for NEON.

\begin{figure}[H]
\includegraphics[width=\textwidth]{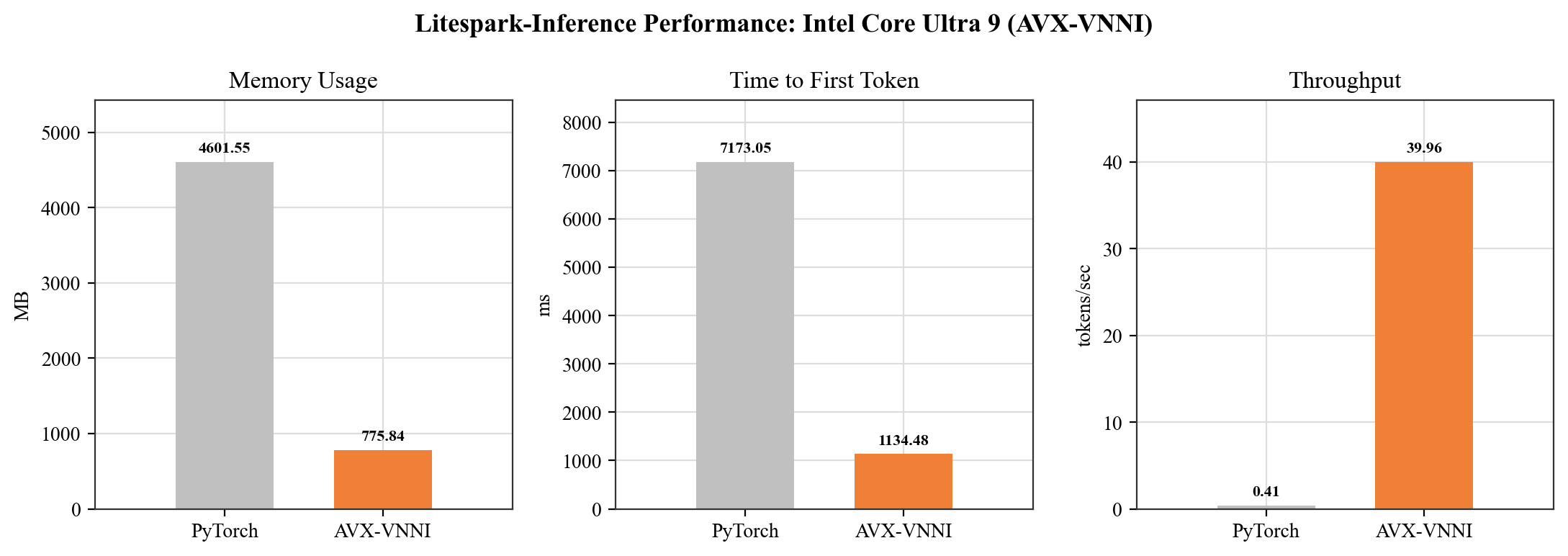}
\caption{Performance comparison on Intel Core Ultra 9 (AVX-VNNI).}
\label{fig:avx_vnni_results}
\end{figure}

\begin{table}[H]
\centering
\begin{tabular}{lccc}
\toprule
\textbf{Metric} & \textbf{PyTorch} & \textbf{AVX-VNNI} & \textbf{Improvement} \\
\midrule
Memory (MB) & 4,601.55 & 775.84 & 5.93$\times$ \\
TTFT (ms) & 7,173.05 & 1,134.48 & 6.32$\times$ \\
Throughput (tok/s) & 0.41 & 39.96 & 97.46$\times$ \\
\bottomrule\\
\end{tabular}
\caption{Results on Intel Core Ultra 9 (AVX-VNNI).}
\label{tab:avx_vnni_results}
\end{table}

The AVX-VNNI results demonstrate that even the 256-bit variant provides substantial speedups over baseline PyTorch. The 39.96 tokens per second throughput is sufficient for interactive use.

\subsection{Cross-Platform Comparison}

\begin{figure}[H]
\centering
\vspace*{0.5cm}
\includegraphics[width=\textwidth]{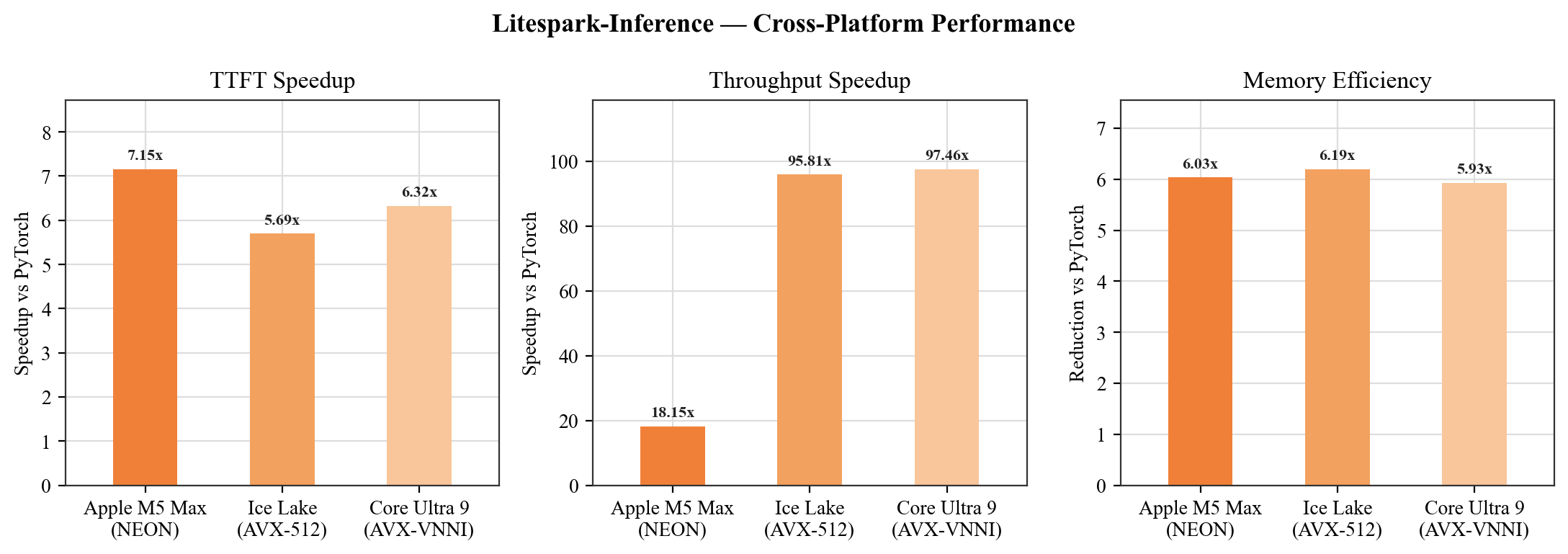}
\caption{Cross-platform performance comparison showing TTFT, throughput, and memory ratios across Apple Silicon, Intel, and AMD processors.}
\label{fig:cross_platform}
\end{figure}

Figure \ref{fig:cross_platform} compares speedups across all tested platforms, normalizing each metric against the PyTorch baseline.

Several key observations emerge from the cross-platform comparison. All platforms achieve substantial throughput improvement over the PyTorch baseline, ranging from 18.15$\times$ on Apple M5 Max (NEON) to 97.46$\times$ on Core Ultra 9 (AVX-VNNI). Memory reduction remains consistent at approximately 6$\times$ across platforms, as this is determined by the weight representation rather than the kernel implementation. All optimized backends improve TTFT, with speedups ranging from 5.69$\times$ on Ice Lake (AVX-512) to 7.15$\times$ on Apple M5 Max (NEON).

\subsection{Energy Consumption Comparison}

\begin{figure}[H]
\centering
\vspace*{0.5cm}
\includegraphics[width=\textwidth]{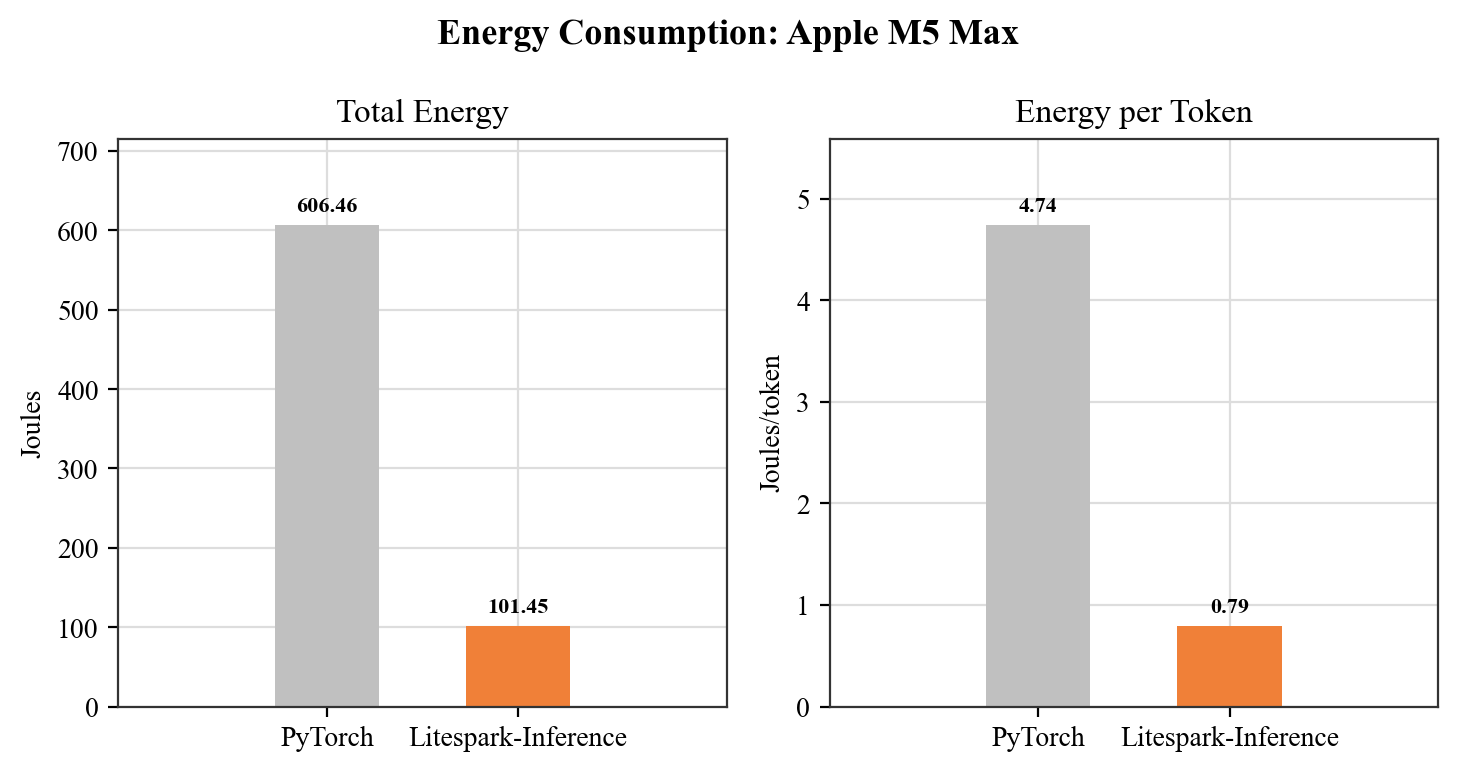}
\caption{Energy consumption comparison on Apple M5 Max.}
\label{fig:power_comparison_apple_m5}
\end{figure}

\begin{figure}[H]
\centering
\vspace*{0.5cm}
\includegraphics[width=\textwidth]{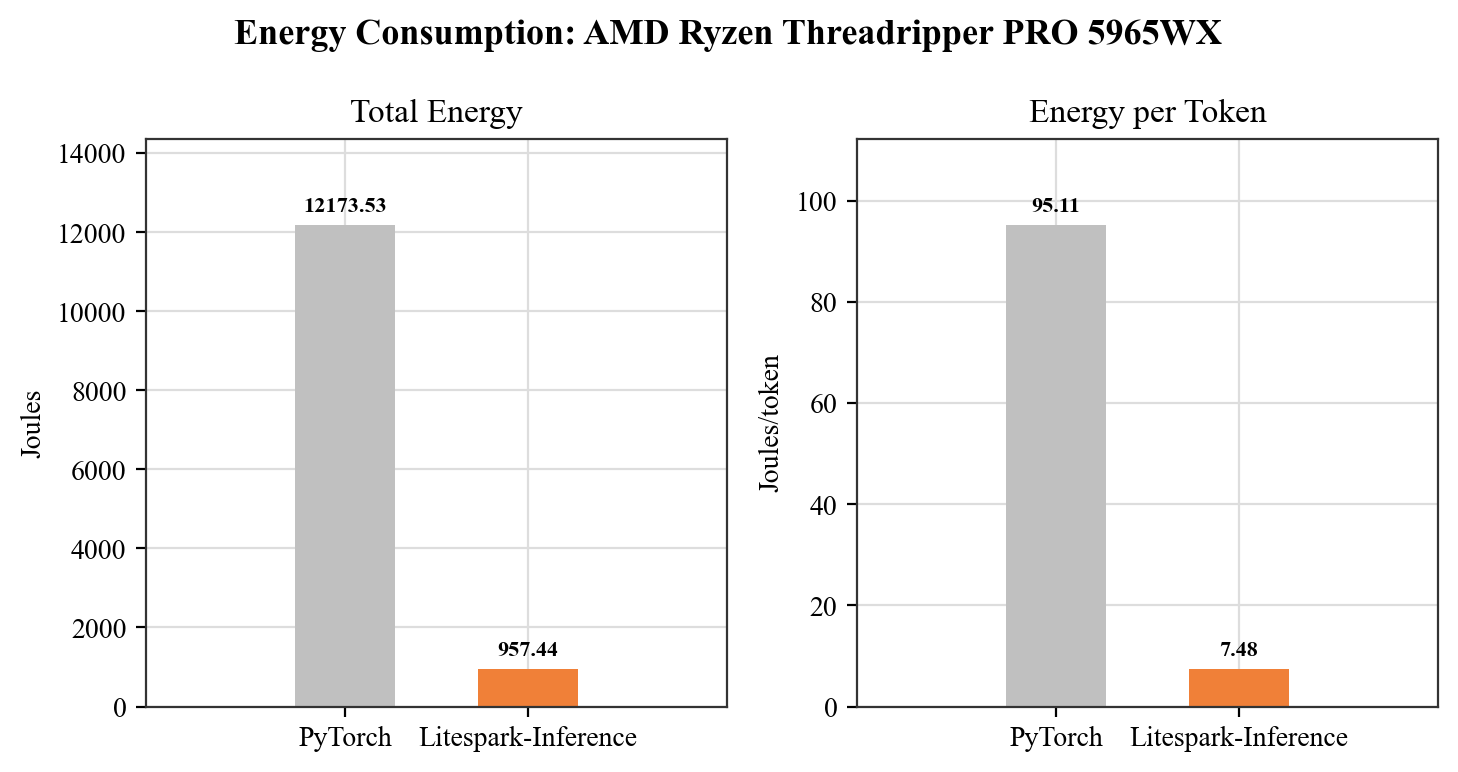}
\caption{Energy consumption comparison on AMD Ryzen Threadripper PRO 5965WX.}
\label{fig:power_comparison_amd}
\end{figure}

Beyond latency and memory usage, local inference must also be energy efficient, particularly for laptops, workstations, and edge devices where sustained power draw affects battery life, thermals, and fan noise. Figures \ref{fig:power_comparison_apple_m5} and \ref{fig:power_comparison_amd} show energy consumption during tg128 generation on Apple M5 Max and AMD Ryzen Threadripper PRO 5965WX, respectively. We measure only the token generation period. On Apple Silicon, CPU power samples from macOS powermetrics are averaged and multiplied by the generation time. On Linux systems with exposed hardware energy counters, package energy is computed from the difference between pre-generation and post-generation counter readings. Energy per token is then calculated by dividing total measured energy by the 128 generated tokens.

\begin{table}[h]
\centering
\small
\begin{tabular}{lcccc}
\toprule
\textbf{System} & \textbf{Metric} & \textbf{PyTorch} & \textbf{Litespark} & \textbf{Improvement} \\
\midrule
\multirow{2}{*}{Apple} 
& Total energy (J) & 606.46 & 101.45 & 5.98$\times$ \\
& Energy/token (J) & 4.74 & 0.79 & 5.98$\times$ \\
\midrule
\multirow{2}{*}{AMD} 
& Total energy (J) & 12,173.53 & 957.44 & 12.71$\times$ \\
& Energy/token (J) & 95.11 & 7.48 & 12.71$\times$ \\
\bottomrule \\
\end{tabular}
\caption{Energy consumption during tg128 generation.}
\label{tab:energy_tg128}
\end{table}

Litespark-Inference substantially reduces generation energy on both platforms, as summarized in Table~\ref{tab:energy_tg128}. On Apple M5 Max, total energy drops from 606.46 J with PyTorch to 101.45 J with Litespark-Inference, corresponding to a 5.98$\times$ reduction. Energy per token falls from 4.74 J/token to 0.79 J/token. On the AMD Threadripper PRO 5965WX, the reduction is even larger: total energy drops from 12,173.53 J to 957.44 J, while energy per token falls from 95.11 J/token to 7.48 J/token, corresponding to a 12.71$\times$ improvement. These gains follow from the same design choices that improve throughput: compact ternary weights reduce memory traffic, and SIMD integer dot-product kernels replace dense floating-point matrix multiplication. Since token generation is both faster and more energy efficient, the total energy required per generated token decreases sharply.

For scale, a 1 kW appliance such as a typical microwave oven consumes 1,000 J per second. On the AMD Threadripper system, PyTorch's tg128 run therefore uses energy comparable to approximately 12 seconds of microwave operation, while Litespark-Inference uses approximately 1 second.

\section{Discussion}

The experimental results demonstrate substantial performance improvements across all tested platforms.
We now analyze the sources of these gains, discuss practical considerations for deployment, and
identify directions for future work.

\subsection{Why Such Large Speedups?}

The 18.15--97.46$\times$ throughput improvements we observe are not the result of a single optimization, but rather the compounding effect of multiple factors working together.

First, ternary weights eliminate floating-point multiplication entirely. Standard matrix multiplication requires expensive floating-point multiply-accumulate operations. With ternary weights, multiplication becomes trivial: multiplying by $+1$ is a no-op, multiplying by $-1$ is negation, and multiplying by $0$ is skipping. This fundamentally changes the computational bottleneck from arithmetic to memory bandwidth.

Second, int8 representation enables direct use of hardware dot product instructions. Modern CPUs include specialized SIMD instructions (SDOT, VPDPBUSD) designed specifically for neural network inference. These instructions process 16--64 int8 pairs per cycle with high throughput. By storing weights as int8, we can feed data directly to these instructions without unpacking or conversion.

Third, the approximately 6$\times$ memory reduction improves cache utilization. A 2B parameter model requires 8 GB and 4 GB in float32 and float16 respectively, but only 0.78--0.81 GB in our int8 ternary format. This allows the entire model to fit in CPU cache hierarchies more effectively, reducing memory bandwidth pressure. When frequently accessed weights are served from cache rather than DRAM, access latency drops from hundreds of cycles to tens of cycles.

Fourth, the KV cache optimization reduces redundant computation during autoregressive generation. Without caching, each new token would require recomputing attention over the entire sequence. The cache converts this $O(n^2)$ cost into $O(n)$, providing speedups that grow with sequence length.

The baseline PyTorch implementation suffers from treating ternary weights as dense floating-point matrices. It performs full floating-point matrix multiplication, uses approximately 6$\times$ more memory, and cannot leverage specialized integer dot product instructions. Our implementation exploits the ternary structure at every level of the system.

\subsection{Accuracy Considerations}

A natural concern with int8 quantization is whether it degrades generation quality. We address this through careful measurement and comparison.

We quantify numerical deviation using the \textit{maximum logit difference}: the largest absolute difference between Litespark-Inference's int8 output and the float32 reference across all logits. For the NEON kernel on Apple Silicon, this difference is approximately 0.68, the largest deviation in any logit value is less than 1.0.

In practice, this small numerical difference has no observable effect on generation quality. Language model generation uses sampling or greedy decoding based on logit rankings. A deviation of 0.68 is insufficient to change which token has the highest probability in almost all cases. We verified this by generating hundreds of samples with both the int8 and float32 implementations: the outputs are identical.

\subsection{Limitations}

Our implementation targets ternary models specifically. The techniques do not directly apply to models using 4-bit or 8-bit quantization, which require different kernel designs. However, the general principle, exploiting quantization structure with hardware SIMD instructions, applies broadly.

The current implementation has been validated on the BitNet architecture. Adapting to other ternary architectures would require modifying the model loading and layer replacement logic, though the core kernels would remain unchanged.

Performance is limited by memory bandwidth rather than compute throughput. On all tested platforms, the kernels are memory-bound: they spend more time loading weights from memory than computing dot products. Future work could explore weight compression or caching strategies to further reduce bandwidth pressure.

\subsection{Future Work}

Several directions could extend this work:

\textbf{Mobile deployment.} ARM Holdings' NEON architecture, with its SDOT instruction, is deployed across billions of mobile devices worldwide, from Apple's iPhone and iPad lineup to Android smartphones and tablets powered by Qualcomm, Samsung, and MediaTek processors. The kernels developed for Apple Silicon use the same ARM instruction set available on these mobile platforms. Adapting the implementation for iOS and Android would enable on-device LLM inference without network connectivity, leveraging ARM's global ecosystem. This would be particularly impactful given ARM's dominant position in mobile computing, where power efficiency and on-device processing are critical. ARM's consistent architecture across their product lines, from high-performance processors in Apple's M-series chips to energy-efficient cores in mobile SoCs, means that optimizations developed for one ARM platform can benefit the entire ARM ecosystem.

\textbf{Larger models.} Ternary training has been demonstrated at 2B parameters. As larger ternary models become available (10B, 70B+), our kernels would scale naturally, with memory savings becoming even more critical.

\textbf{Training support.} Current work focuses on inference. Extending the kernels to support backward passes would enable efficient fine-tuning of ternary models on consumer hardware.

\textbf{Specialized ternary models.} Litespark-Inference currently serves existing ternary models. Future work could explore training efficient ternary models for specialized domains, including vision and multimodal workloads, extending the benefits of CPU efficient inference beyond general purpose text generation.

\textbf{Hybrid quantization.} Some models use ternary weights with higher-precision activations (e.g., int16 or bfloat16). Adapting the kernels to support mixed-precision computation could improve accuracy while maintaining most of the performance benefits.

\textbf{Agentic workloads.} Agentic systems make repeated model calls, and multi-agent workflows amplify this cost as multiple local agents coordinate across execution. The high throughput, low latency token generation achieved by Litespark-Inference makes local CPU inference practical for ternary models in such workloads. This is especially useful for privacy, cost reduction, and offline execution for edge use-cases.

\subsection{Extending to Non-Ternary Models}

While our kernels target ternary weights, the underlying techniques generalize to other quantization schemes. The key insight is matching the quantization format to available hardware instructions.

For 4-bit quantization, one could pack two 4-bit weights into each int8 value and use similar SIMD dot product instructions. For 2-bit quantization, four weights could be packed per byte. The challenge is balancing the cost of unpacking against the benefit of reduced memory bandwidth.

For asymmetric quantization (where zero-point is non-zero), the zero-point correction logic would need adjustment, but the core SIMD computation remains the same.

The broader lesson is that hardware-aware quantization, choosing quantization formats that align with available instructions, can unlock substantial performance improvements on consumer CPUs.

\section{Conclusion}

This work demonstrates that ternary neural networks can achieve practical, high-performance inference on consumer CPUs through hardware-aware kernel design. By exploiting the structure of ternary weights and leveraging modern SIMD dot product instructions, we achieve 18.15--97.46$\times$ throughput improvements and approximately 6$\times$ memory reduction compared to standard PyTorch implementations in the case of the BitNet 2B model.

The key insight is that extreme quantization, restricting weights to just three values, enables a qualitatively different computational approach. Rather than treating quantization as an approximation to full-precision computation, we design kernels specifically for the ternary case, eliminating floating-point multiplication entirely and using specialized integer instructions.

Litespark-Inference packages these optimizations as a pip-installable library with automatic platform detection and HuggingFace integration, making efficient ternary inference accessible to developers without specialized knowledge of SIMD programming or hardware architecture.

The results suggest that ternary models like BitNet represent a promising direction for democratizing large language model deployment. A 2-billion parameter model that fits in approximately 0.78--0.81 GB memory and generates up to 41 tokens per second on consumer hardware enables applications previously requiring expensive GPU infrastructure or cloud API calls.

As ternary training techniques continue to improve and larger models become available, the performance benefits demonstrated here will become increasingly important. The ability to run billion-parameter models efficiently on consumer CPUs, from desktop workstations to laptops to mobile devices, could fundamentally change how we think about deploying and using large language models.

\bibliographystyle{plain}

\appendix

\section{Comparison with BitNet.cpp v2}
\label{appendix:bitnetcpp}

Shortly after we completed Litespark-Inference, Microsoft released BitNet.cpp v2 \cite{bitnetcppv2}, corresponding to the 01/15/2026 BitNet CPU Inference Optimization update. This update introduced additional CPU inference optimizations, including improved low-bit kernels and optimized execution paths for BitNet models. We benchmarked both implementations on identical hardware using batch size 1 across all platforms and all framework versions to compare their performance characteristics.

\subsection{Benchmark Setup}

We followed Microsoft's pp128+tg128 methodology using their \texttt{e2e\_benchmark.py} script: processing a 128-token prompt (pp128, measuring prefill speed) followed by generating 128 new tokens (tg128, measuring autoregressive generation speed). All tests used the BitNet b1.58 2B-4T model. For thread-scaling measurements, we swept 1, 2, 4, 8, 10, and 16 CPU threads.

We tested on three platforms:
\begin{itemize}
    \item \textbf{AMD EPYC 9R14} (AWS c7a.4xlarge): 16 vCPUs, AVX-512 VNNI
    \item \textbf{Intel Xeon Platinum 8488C} (AWS c7i.4xlarge): 16 vCPUs, AVX-512 VNNI
    \item \textbf{Apple M5 Max} (MacBook Pro): 10-core CPU, ARM NEON with SDOT
\end{itemize}

\subsection{x86 Results}

\begin{figure}[H]
\centering
\includegraphics[width=\textwidth]{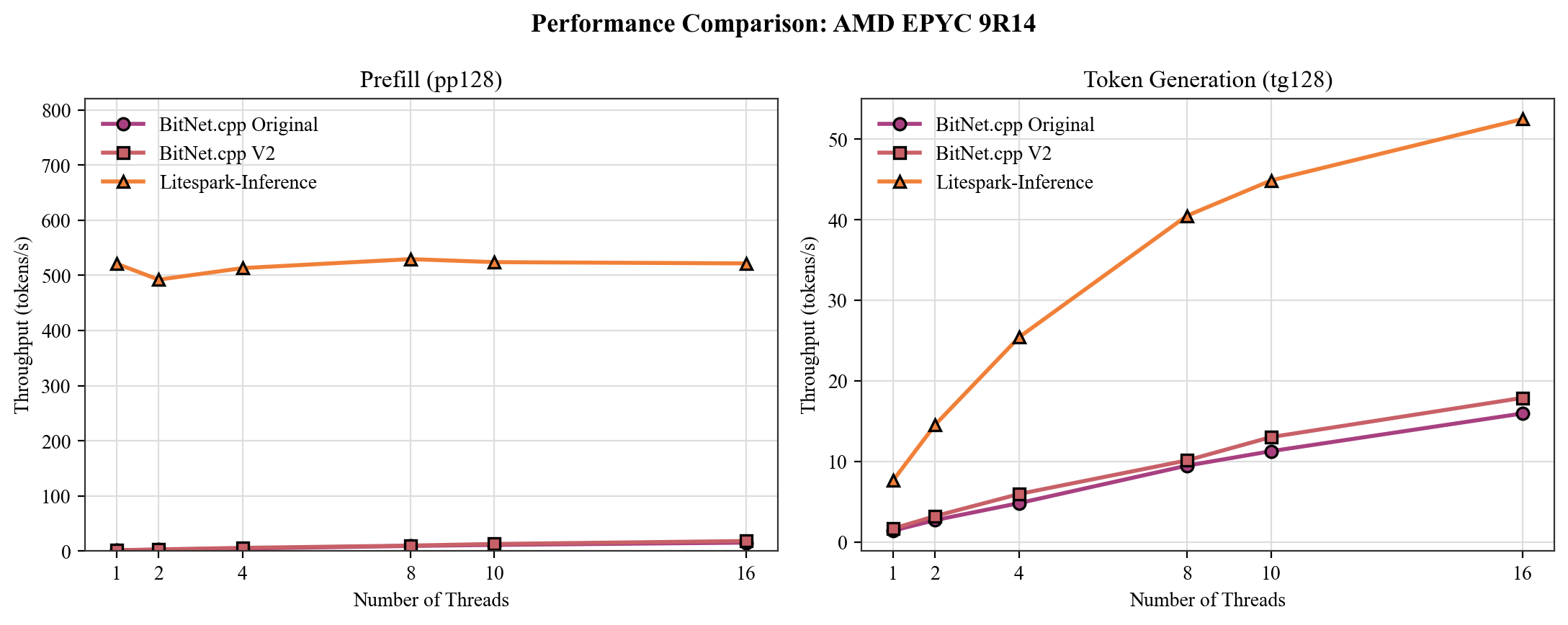}
\caption{Scaling behavior on AMD EPYC 9R14 across BitNet.cpp Original, BitNet.cpp V2, and Litespark-Inference.}
\label{fig:amd_v2}
\end{figure}

\begin{figure}[H]
\centering
\includegraphics[width=\textwidth]{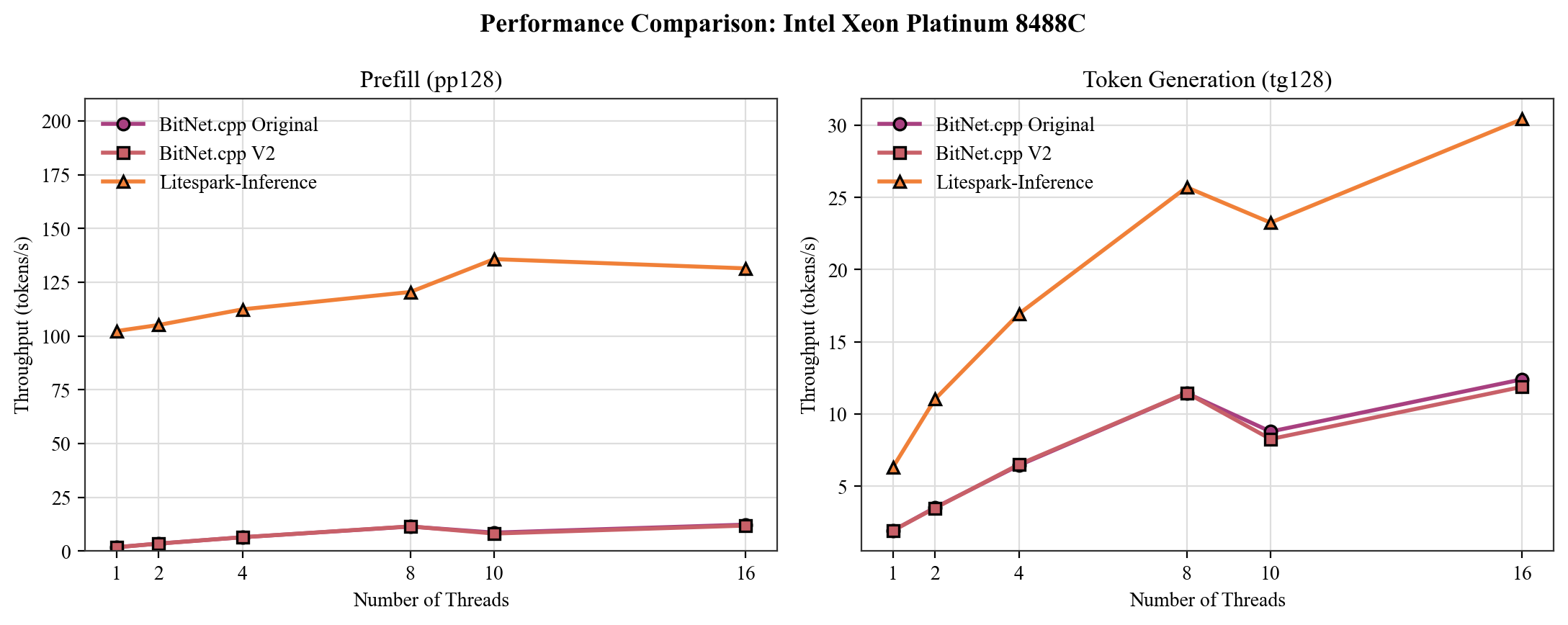}
\caption{Scaling behavior on Intel Xeon Platinum 8488C across BitNet.cpp Original, BitNet.cpp V2, and Litespark-Inference.}
\label{fig:intel_v2}
\end{figure}

Figures \ref{fig:amd_v2} and \ref{fig:intel_v2} visualize how performance scales with thread count on the x86 platforms listed above.

\subsection{ARM Results (Apple Silicon)}

\begin{figure}[H]
\centering
\includegraphics[width=\textwidth]{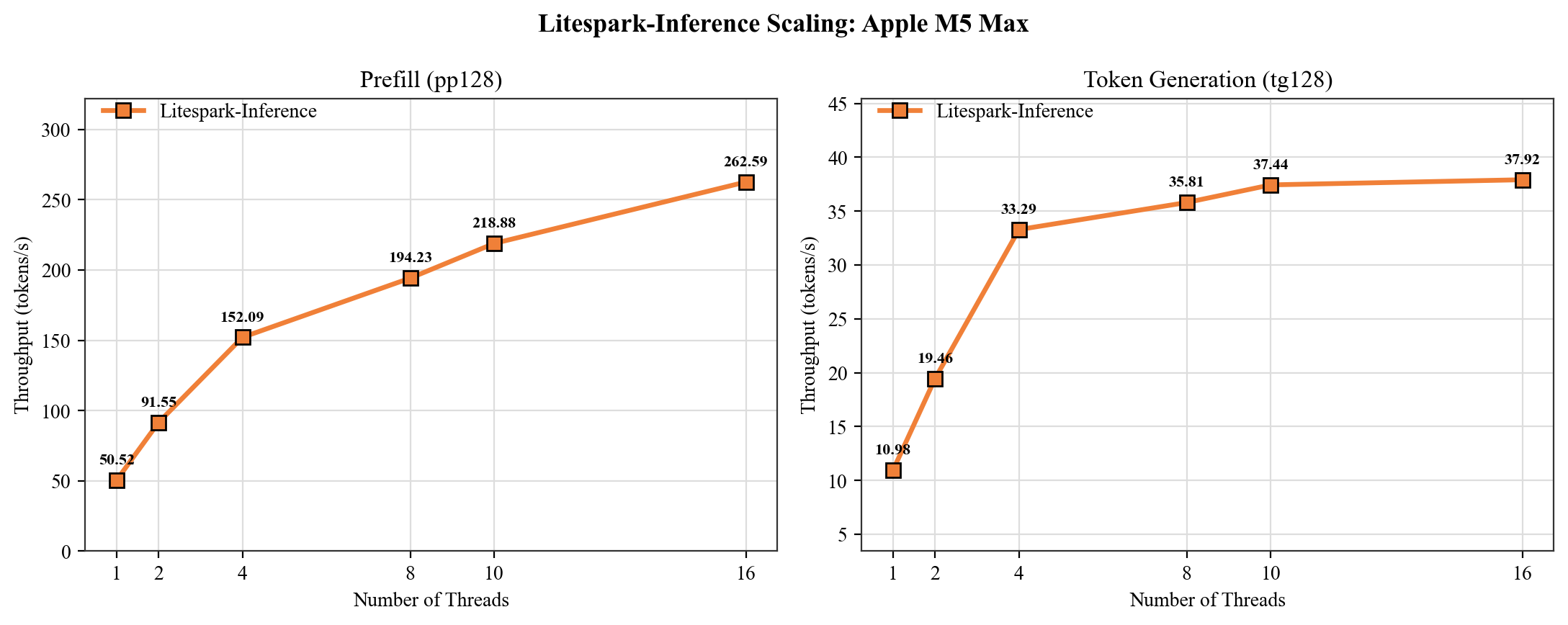}
\caption{Litespark-Inference scaling on Apple M5 Max.}
\label{fig:m4_scaling}
\end{figure}

Table \ref{tab:m4_scaling} and Figure \ref{fig:m4_scaling} show Litespark-Inference's performance scaling on Apple M5 Max. These results use the same pp128+tg128 measurement protocol and the NEON SDOT kernel.

\begin{table}[H]
\centering
\begin{tabular}{lcc}
\toprule
\textbf{Threads} & \textbf{Prefill pp128 (tok/s)} & \textbf{Generation tg128 (tok/s)} \\
\midrule
1 & 50.52 & 10.98 \\
2 & 91.55 & 19.46 \\
4 & 152.09 & 33.29 \\
8 & 194.23 & 35.81 \\
10 & 218.88 & 37.44 \\
16 & 262.59 & 37.92 \\
\bottomrule\\
\end{tabular}
\caption{Litespark-Inference performance on Apple M5 Max using the pp128+tg128 protocol.}
\label{tab:m4_scaling}
\end{table}

\subsection{MacBook Neo Support}

MacBook Neo is a recently introduced A18 Pro based MacBook in Apple's current hardware lineup, designed around everyday productivity, education workflows, on-device AI features, and long battery life \cite{applemacbookneonewsroom, applemacbookneospecs}. We evaluated Litespark-Inference on this system to verify that the same NEON SDOT backend used on M-series Macs also runs efficiently on Apple's A-series Mac hardware.

\begin{figure}[H]
\centering
\includegraphics[width=\textwidth]{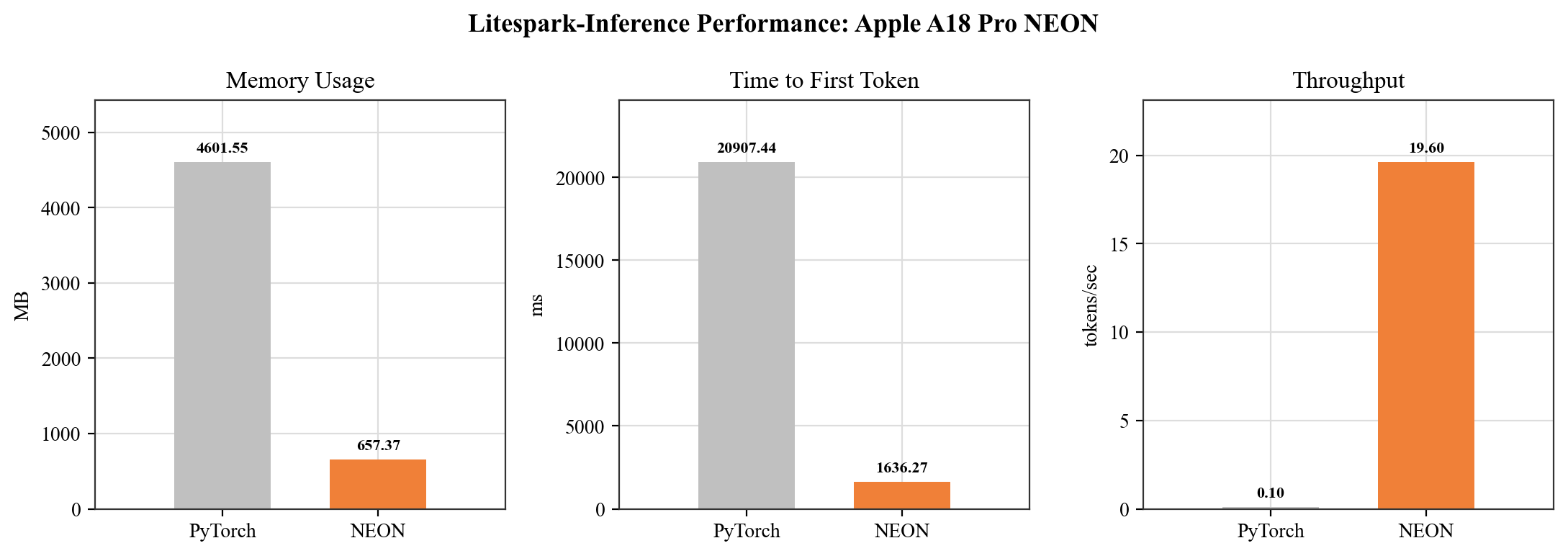}
\caption{Performance comparison on MacBook Neo using Apple A18 Pro.}
\label{fig:neo_appendix}
\end{figure}

\begin{table}[H]
\centering
\begin{tabular}{lccc}
\toprule
\textbf{Metric} & \textbf{PyTorch} & \textbf{NEON} & \textbf{Improvement} \\
\midrule
Memory (MB) & 4,601.55 & 657.37 & 7.00$\times$ \\
TTFT (ms) & 20,907.44 & 1,636.27 & 12.78$\times$ \\
Throughput (tok/s) & 0.10 & 19.60 & 196.00$\times$ \\
\bottomrule\\
\end{tabular}
\caption{MacBook Neo results on Apple A18 Pro.}
\label{tab:neo_appendix}
\end{table}

Together with the M-series Apple Silicon, Intel, and AMD results, this evaluation shows that Litespark-Inference supports the CPU execution paths evaluated in this paper across high-performance systems and new Apple Silicon hardware.

\subsection{Apple Silicon Accelerate Support}

Litespark-Inference also supports an Apple Accelerate path for workflows that require float32 computation. This path uses Apple's Accelerate framework, which can leverage the Apple Matrix Extensions (AMX) coprocessor \cite{appleamx}. We evaluated this path on Apple M5 Max using the same model and reporting the same memory, TTFT, and throughput metrics used in the main Apple Silicon experiment.

\begin{figure}[H]
\centering
\includegraphics[width=\textwidth]{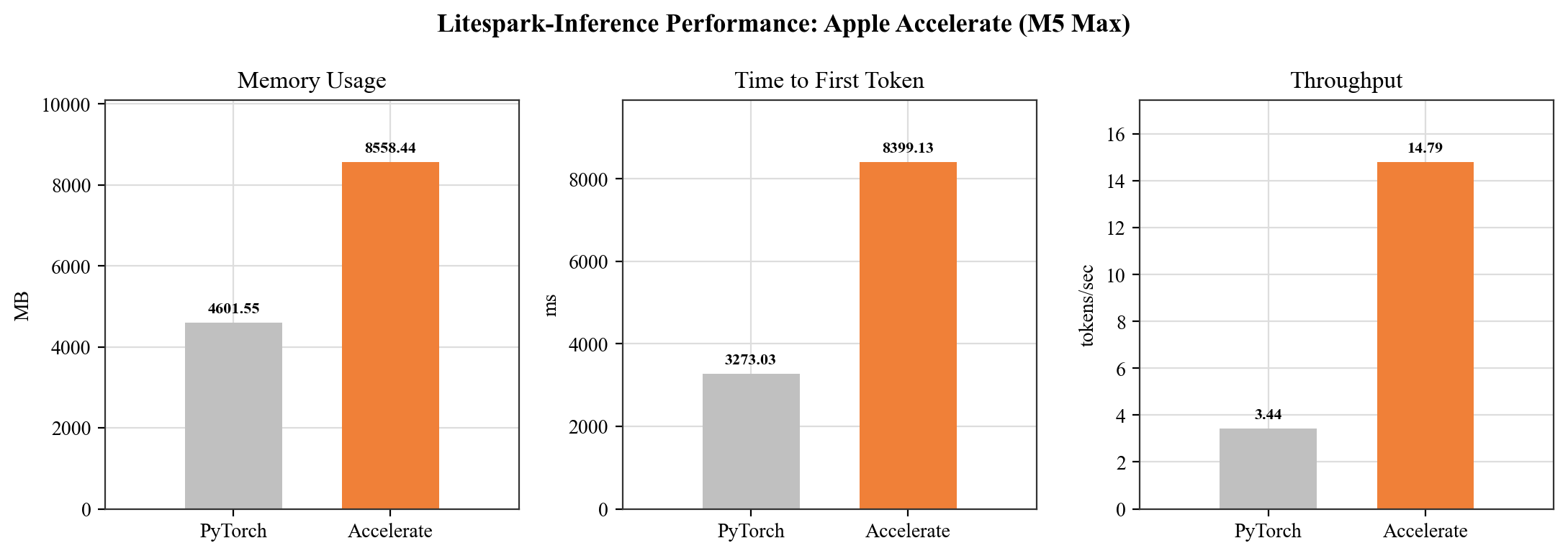}
\caption{Apple Accelerate support results on Apple M5 Max.}
\label{fig:apple_accelerate_appendix}
\end{figure}

\begin{table}[H]
\centering
\begin{tabular}{lccc}
\toprule
\textbf{Metric} & \textbf{PyTorch} & \textbf{Accelerate} & \textbf{Relative result} \\
\midrule
Memory (MB) & 4,601.55 & 8,558.44 & 1.86$\times$ higher \\
TTFT (ms) & 3,273.03 & 8,399.13 & 2.57$\times$ slower \\
Throughput (tok/s) & 3.44 & 14.79 & 4.30$\times$ higher \\
\bottomrule\\
\end{tabular}
\caption{Apple Accelerate support results on Apple M5 Max. Memory is peak RSS, TTFT is time to first token, and throughput is tokens per second during generation.}
\label{tab:apple_accelerate_appendix}
\end{table}

\end{document}